\documentclass[twoside,11pt]{article}

\usepackage{jmlr2e}

\usepackage{here}
\usepackage[usenames,dvipsnames,svgnames,table]{xcolor}
\usepackage{hyperref}

\jmlrheading{1}{2016}{x}{xx}{xxx}{Lambert Schomaker}

\ShortHeadings{Caveats on Bayesian and hidden-Markov models}{L. Schomaker}
\firstpageno{1}

\begin{document}


\title{Caveats on Bayesian and hidden-Markov models - {\color{lightgray} v2.8}}

\author{\name Lambert Schomaker \email schomaker@ai.rug.nl \\
       \addr Artificial Intelligence \& Cognitive Engineering Institute \\
       University of Groningen\\
       Nijenborgh 9, NL-9747 AG, The Netherlands}
       

\editor{~}

\maketitle

\begin{abstract}
This paper\footnote{This paper has been on shelf for some time,
since 2009 and is the result of the extensive discussions with
at-the-time MSc student Jean-Paul van Oosten, colleagues in handwriting
recognition and critical/supportive comments by Bob Duin.} 
describes a number of fundamental and practical
problems in the application of hidden-Markov models and Bayes
when applied to cursive-script recognition. Several problems,
however, will have an effect in other application areas.
The most fundamental problem is the propagation of error in the 
product of probabilities. This is a common and pervasive problem
which deserves more attention. On the basis of Monte Carlo modeling,
tables for the expected relative error are given. It seems that
it is distributed according to a continuous Poisson distribution over
log probabilities. A second
essential problem is related to the appropriateness of the Markov assumption.
Basic tests will reveal whether a problem requires modeling
of the stochasticity of seriality, at all. Examples are given of lexical
encodings which cover 95-99\% classification accuracy of a lexicon, 
with removed sequence information, for several European languages. 
Finally, a summary of results on a non-Bayes, non-Markov method in
handwriting recognition are presented, with very acceptable results
and minimal modeling or training requirements using nearest-mean classification.
\end{abstract}


\section{Introduction}

The use of Bayes and hidden-Markov models (HMMs) is pervasive in machine
learning.  The Bayes' Theorem~\citep{Bayes} describes the theoretically correct
optimum for the posterior probability given observations.  The
theoretical soundness of the paradigm makes it seem inappropriate to
pose critical questions.  Similarly, in the application domains of
speech and handwriting recognition, the hidden-Markov model -which is
also Bayesian in its essential computations- is the predominant
paradigm, without essential objections or reserve.  However, both in handwriting
and speech recognition, current word-classification performances are
still not up to the expectations the were prevalent during the last quarter
of the previous century. This is in stark contrast with the dissemination
of other technologies. The invention of the motorized
airplane in 1903 produced a large and flourishing industry within ten years.
This is contrary to the situation in hidden-Markov modeling for handwriting 
and speech recognition. At the same time, neural networks
can be applied easily today by master students in a wide range of
pattern-recognition problems. New paradigms for classification of stochastic time
series, such as the (bidirectional) long-term short-term memory neural networks~\citep{LSTMfirst1997,BLSTM2005} 
are enjoying considerable interest and success. On the contrary, 
a PhD student starting to 
apply HMMs in handwriting recognition should still be prepared for a bumpy ride, 
more than 25 years after the introduction of the method. Essential portions of the {\em know how} 
in the application of HMMs are not spelled out in the scientific literature.
As a consequence, the reported successes from literature cannot be replicated easily.
Therefore, time is ripe for critical considerations. Elsewhere~\citep{vanOostenMarkov2014,vanOosten2017}, we have addressed problems in estimating hidden-Markov models, as evidenced from
haphazard model-fit value trajectories during regular Baum-Welch training.

\medskip

In a similar vein, although Bayes' rule is optimal, the same cannot be said 
about naive-Bayes classifiers or Bayesian models in many current application domains. 
Despite a strong advocacy and a steep increase in Bayesian studies, it is often
difficult to replicate successes of others. What is more intriguing, however, is that
non-Bayesian pattern classifiers and systems yield competitive or improved performances using neural networks or even nearest-neighbor classifiers,
with limited (human) development effort. It is time to take a step back.

\medskip 

First, a number of
fundamental problems with Bayes will be introduced.  Then, 
problems with hidden Markov models will be laid out. In the final
section a case will be made for geometric methods: distance or
dissimilarity-based vectorial matching, which remain the most convenient
approach for conditions with limited ground-truth reference data.

\section{Problems with a Scotsman}

The Scottish reverend Thomas Bayes~\cite{Bayes} introduced the well-known
equation for estimating the probability of a fact $C$ given an
observable $X$

\begin{equation} \label{eq:Bayes}
   P(C|X) = \frac{P(X|C) * P(C)}{P(X)} 
\end{equation}

where $P(C)$ is the prior probability of $C$ occurring, at all,
$P(X)$ is the prior probability of $X$ occurring, at all,
while $P(X|C)$ is the probability of observing $X$ under the
condition that $C$ is known to be true. 
In pattern recognition, $X$ will coincide
with the observed feature value(s), while $C$ will coincide with
a particular class of which the presence must be estimated.
The Bayes theorem describes the benefits 
of weighing the probability of $X$ under condition of $C$ with
two factors:

\begin{itemize}
\item Multiplication with $P(C)$ represents realism ("how often does C occur, anyway?"); 
\item Division by $P(X)$ exploits informational value ("if the general
probability of $X$ occurring is low, observing $X$ tells a lot"). 
\end{itemize}

Application of the Bayes equation will yield a clarifying (contrast-enhancing) 
scaling operator on the {\em a-posteriori} probability $P(C|X)$, such that in 
a multi-class problem the class $C_i$ with the largest a-posteriori probability can 
be easily identified. The use of probabilities as the currency in reasoning or in
'belief propagation' is definitely an advantage over methods which use
more diversely scaled measures of similarity or rankings, e.g., for combining
classifiers~\cite{Erp}. However, 
combination of evidence from multiple sources is still not trivial.
In order to combine pieces of information, and disregarding 
complex interactions due to stochastic interdependence of 
observables, a so-called naive Bayes scheme is used, usually. The
effect of a conjoint observation of a set of observables $X = {x_1,x_2,...,x_N}$ is 
then inserted as a product of probabilities:

\begin{equation}
   P(C|x_1,x_2,...,x_N) = P(C) * \prod_{j=1}^N \frac{P(x_j|C)}{P(x_j)} 
\end{equation}

It may not be appropriate to combine evidence by assuming conjunction of stochastically
independent observations. This shortcoming of the naive-Bayes assumption
has been dealt with in depth elsewhere~\citep{Rennie2003}.
Note that the computation of a {\em product of probabilities}
is essential for a classifier which is implemented as a Bayesian engine.
This is made more explicit in the following formulation, demonstrating the
presence of two such product chains:

\begin{equation}
   P(C|x_1,x_2,...,x_N) = P(C) * \frac{ \prod_{j=1}^N P(x_j|C)}{\prod_{j=1}^N P(x_j)} 
\end{equation}

We will get to this aspect in a later section. For now,
a basic question is whether learning systems should be implemented
as a Bayesian system, by using the above formalism. It should be realized
that a classifier may operate close to the optimal Bayes performance,
without being implemented as a Bayes engine, using estimations of probability,
etc. A well-known example is the fact that a nearest-neighbor classifier with
an infinite number of examples will yield an error which is guaranteed to be less
than twice the optimum Bayes error~\cite{Duda73}. In fact, a Bayesian implementation
of a classification system may be impractical in a realistic system, as will be
illustrated in the following example.

\noindent 
\begin{example}
Let us translate
the problem to an ecologically realistic problem which is presented to,
e.g., an ancestral primate. The subject, an autonomous intelligent system,
is rather young, and aims at survival. It encounters a hissing
and rattling snake and would like to estimate whether a bite
is to be expected, i.e, in Bayesian terms:
\end{example}

\begin{equation}
   P(bite|snake) = \frac{P(snake|bite) * P(bite)}{P(snake)}
\end{equation}

For a learning system, implemented in the manner of many contemporary
machine-learning models, this would imply the following
instructions to the subject in order to compute the necessary terms on 
the right-hand side:

\begin{description}

\item[$P(snake)$] walk on the earth extensively, count the instances for
all classes of objects you encounter and if you have seen enough, jot
down the probability of encountering snakes;

\item[$P(bite)$] interact with object classes as extensively as
possible, and count the number of times you were bitten;

\item[$P(snake|bite)$] each time you are bitten, keep count of the
number of times the perpetrator was a snake. 

\item[$Note:$] Make sure you encounter sufficient numbers of snakes,
bites, and biting snakes, as you want those probabilities to be
reliable;

\item[$P(bite|snake)$] can now be computed.

\end{description}

While this is a colloquial, non-formal example it illustrates a number
of underlying problems with the Bayesian formalism, even for
single-observation problems.  To obtain reliable priors, huge amounts of
data need to be collected, that a newborn system is unlikely to possess. 
This makes a classification system which is {\em implemented by means
of} the Bayes equation rather impractical.  For instance, in natural cognitive
systems, the phenomenon of single-shot (one-shot, or one-trial) 
learning~\citep{ GuthrieOneTrial1948,UnderwoodKeppelOneShot1962,
HerbstDissSingleShot1982,HintonTechRepSingleShot1984,SparseOneShotSussman1997}
will make extensive and
painful sampling unnecessary, after the first bite.  It is even likely
that the original perpetrator of the first bite does not need to to be a
snake to estimate that a bite is imminent when seeing an instance of the
hitherto unseen species ``snake''.  For the problem at hand, we can
largely deconstruct the Bayes equation:

\medskip 

Since a concrete version of Example 1 is concerned with estimating the
likelihood of a number of possible snake behaviors, one of which is
biting, say, $\{bite,rattle,misc\}$, then the total problem will be
described by an estimation of three a-posteriori probabilities:

\begin{eqnarray*}
   P(bite|snake) = P(snake|bite) * P(bite) & / & P(snake) \\
   P(rattle|snake) = P(snake|rattle) * P(rattle) & / & P(snake) \\
   P(misc|snake) = P(snake|misc) * P(misc) & / & P(snake) 
\end{eqnarray*}

It can be seen quickly that the denominator $P(snake)$ does not need 
to be known to find the most likely behavior in this set.
As a result, the agent only has to compute three relative posterior
estimates in the example, denoted as $\tilde{p}$:

\begin{eqnarray*}
   \tilde{p}(bite|snake) = P(snake|bite) & * & P(bite) \\
   \tilde{p}(rattle|snake) = P(snake|rattle) & * & P(rattle) \\
   \tilde{p}(misc|snake) = P(snake|misc) & * & P(misc) 
\end{eqnarray*}

Furthermore, however, we need to take a step back and ask ourselves the
question why we actually need to compute an exact $P(bite|snake)$ at
all.  For an autonomous system, the more fundamental variable is the
utility or cost involved, i.e., the estimated utility $u' =
P(bite|snake) * U(bite)$ where $U(bite)$ is the utility of being bitten. 
Since the cost of dying can be considered infinite, any non-zero
probability of suffering a bite would necessitate a swift avoidance of
the snake, no matter the height of the probabilities $P(rattle)$
and $P(misc)$.  This leaves the original Bayes equation completely
deconstructed, with only the term $P(snake|bite)$ remaining.  The latter
can be translated into a probabilistic question: "is it possible that a
proportion of the bites suffered comes from a snake?".  Note, however,
that we are dealing with an infinite-cost problem, such that it is
sufficient to express the problem as a logical question: "is
it true that $\exists x \; bites(x) \wedge x=snake$?", yielding a value
of one or zero, exempt of probabilistic subtleties.  The exact value of
the probability is becoming less and less important for large costs. 
Under these conditions, the autonomous agent only needs to know whether
the posterior probability equals zero or not and little is left from
Eq.~\ref{eq:Bayes}.  Utility is a
pervasive and real factor, not an outlandish concept for exclusive use within
operations research or robotics.  Most applications of pattern
recognition have a cost factor attached to their decisions. Consider, as an example,
the cost of a False-Miss decision in automatic signature verification 
for a fraudulent bank transaction. The actual deployment of a particular
algorithm in such a context does not only depend on its sensitivity and 
specificity (probabilities) but also on the risk the embedding system 
is able to take (utility).

\medskip 

Even without an excursion into the cost domain, the importance of the
exact value of the prior can be considered as relative. It provides a
pragmatic weighing according to the probability of events. However, in
many problems, such as data mining or knowledge discovery, the pattern
classification effort is exactly taken in order to estimate the occurrence
probability of patterns. In such a case, the prior can be replaced by 
a constant, for instance with a value of one divided by the number of 
classes, and the Bayes equation is deconstructed.

\medskip 

It could be argued that the snake example is extreme, since not all problems
are concerned with infinite costs attached to the decision which is 
taken on the basis of the computed posterior probability. Therefore, four
more examples will be given here, each leading to at least a
partial deconstruction of the Bayes equation.

\medskip 

\noindent
%
%
\begin{example}
A forensic institute maintains a database of handwritten 
fraudulent, black-mail and extortion letters. For an incoming
handwritten document of unknown origin, the posterior probability
is needed for finding the most likely writer in the database. Unfortunately,
the database is contaminated by a single recidivist occupying 90\% of the
records.
\end{example}

Is it desirable to compute the full Bayes posterior?
A judge may rightfully argue that it is more important to know the
likelihood ratio $P(X|C)/P(X)$ for the visual, pattern-recognition evidence 
in isolation and disregard the (pragmatic) probability of encountering a writer
in this particular database, i.e., $P(C)$ in Eq.~\ref{eq:Bayes}.
In the legal discussion about the evidence,
the value of (a) the decision by the pattern-recognition tool after using
pure-shape evidence and (b) the knowledge about the presence of a recidivist 
in the database can and should be decoupled and treated as isolated pieces 
of evidence. In other words, the designer of the pattern-recognition 
tool does not want to suffer a legal suit because of an erroneous system
decision which is not caused by the applied pattern-recognition algorithm 
proper, but by the Bayesian weighing with a prior, for this utilization 
context. A decoupling of the application and context specific prior
$P(C)$ from the actual machine-learning information $P(X|C)/P(X)$ 
is desirable, from this perspective. Given that $P(X)$ would be the
same for all $C$, this leaves us, again, with the deconstructed Bayes
term $P(X|C)$, which is, according to the author the essential contribution
of pattern recognition proper: Knowing under which conditions patterns do
occur. The other two terms are context-dependent biasing factors.

\medskip

%
%
\noindent
\begin{example}
A cultural-heritage institution provides access to a large manuscript
collection of administrative texts. The collection is also interesting to
genealogy enthusiasts tracking their family name. Given a word-image
example of a handwritten family name, the system must retrieve the most similar
images.
\end{example}

Also in this application, the user is not interested in the
clean-Bayesian posteriors, but is focused to find a unique 
needle-in-a-haystack family name and does not want to be annoyed by
the fact that $P(Smith)$ and $P(Johnson)$ are very large values,
causing the ``hit list'' for targets $Smyth$ or $Jonson$ to be contaminated 
with irrelevant items with a larger prior.

\medskip 

%
%
\noindent
\begin{example}
When training a multi-layer perceptron to classify the digits one
to nine, the system can be made to behave Bayesian if it is trained
with the natural~\citep{Benford1938} frequencies of training instances of the
digits: 1 (30.1\%), 2 (17.6\%), 3 (12.5\%), 4 (9.7\%), 5 (7.9\%), 6 (6.7\%),
7 (5.8\%), 8 (5.1\%), 9 (4.6\%). Is this appropriate?
\end{example}

On the contrary, in machine learning, most system designers will ignore 
the Benford distribution 
of digits and use a balanced training set with an equal number of samples for 
each of the classes, 
such that also the low-frequency digit ``9'' will be recognized reliably and
the classifier will provide an output likelihood based on shape evidence
alone, rather than on a mixture of shape evidence and pragmatic knowledge on
class priors. For the sake of argument, the digit zero, which is obviously 
needed in a real system, is left out of this example:
Its prior is excluded from the Benford distribution. 

In patterns of human origin, e.g., spoken~\citep{SpeechVar2007} or
written words~\citep{Wing2000,Tucha2004}, there is an additional
confounding factor.  The low-frequency patterns will be produced in a
less stable manner and with a lower signal quality than the
high-frequency words which have enjoyed more human-motor training.  For
this reason it is better to increase the relative presence of instances from
low-frequency classes in a training set, rather than to rely on Bayes to 
decide for sampling frequencies. Also in latent semantic indexing~\citep{Dumais97automaticcross-linguistic,MikolovWord2Vec2013}, 
effort is spent in non-Bayesian sampling a corpus of text in such a way 
that the model
is {\bf not} determined by the prior probability of terms. High-frequency
function words would dominate deep bag-of-words models to 
the extent of becoming unusable.

%
%
\noindent
\begin{example}
In a criminal case, the forensic expert has computed the probability 
that some features of handwriting (or speech, etc.) in a sample are 
produced by the suspect. In the denominator of the Bayes equation, the 
probability of these features in the general population are used. The suspect,
however, claims to be right-handed, that this is an accepted fact
and demands that the probability is
recomputed for the proper conditional reference set of right-handed writers.
In many countries, the judge is allowed to disregard such 'subtleties'.
However, the example illustrates the arbitrariness of reference sets. Both 
the prior of the decision (class) and the probability of evidence (feature) 
are implicitly conditional to some world context $W_{c}$ and $W_{x}$ where the
observables happen, respectively:

\begin{equation} \label{eq:RealisticBayes}
   P(C|X,W_{c},W_{x}) ~= \frac{P(X,W_{x}|C,W_{c}) * P(C,W_{c})}{P(X,W_{x})}    
\end{equation}

In words: the probability that decision $C$ is true, given observation $X$ is
considered in conjunction with a particular context $W_{c}$ generating the
decisions and a particular context $W_{x}$ generating that observation.
The context determinants $W$ are then factors that should be taken into 
account in computing the conditional probability of observing $X$ given
$C$, in the adjusted context-specific prior for $C$ and the evidence-generating
context for observing $X$. The fact that context $W$ plays a role becomes
evident in any machine learning task where there is a difference between
training and testing due to $W_{test} \ne W_{train}$ such that $P(X,W_{test,x}) \ne P(X,W_{train,x})$, $P(C,W_{test,c}) \ne P(C,W_{train,C})$ and $P(X,W_{test,x}|C,W_{test,c}) \ne P(X,W_{train,x}|C,W_{train,c})$.

Both in legal cases and in a reasoning robot, considerable deliberation over
the validity of the reference context is needed when using a Bayesian
decision framework.

\end{example}

\subsection{Five remarks on Bayes}

To summarize, it can be noted that -in addition to the known problems 
of using naive Bayes- there are five other essential caveats:

\begin{description}
\item[``behave as'' vs ``compute as'' Bayes?] Whereas the Bayes equation
is optimal, learning by estimating its right-hand side terms may be
rather impractical from a 'tabula rasa' condition or only a limited number
of examples. It is very well possible that a working classifier
performs exactly according to Bayes, without ever residing to the
computation of Bayesian products of probabilities 'under the hood' of
its machinery;

\item[probability or utility?] Since for autonomous systems, the
ultimate decision is based on utility, it is conceivable 
to estimate utility directly from observables,
bypassing the cumbersome estimation of probabilities, as is
done in reinforcement learning~\citep{WieringOtterloRL2014};

\item[probability or possibility?] There is a singularity at the
probability of zero, such that an autonomous system cannot know whether
a zero-probability estimate corresponds to a structural, physical impossibility, 
or, alternatively, may be the consequence of insufficient sampling. 
However, both in Bayesian and Markov modeling, such problems are
circumvented by using default, 'back-off' probability constants with
a low value (euphemistically called 'smoothing'). Such a practical
intervention appears to be in conflict with the clear-cut cleanliness
of the Bayes equation and will be a problem for cases where the
estimated zero probability indeed represents a structural impossibility.

\item[cost of learning?] How to obtain reliable priors and class-conditional
feature distributions? The bootstrap problem may be insurmountable.
In robotics problems applying Bayesian learning it is not uncommon to find 
the batteries drained and mechanical defects appearing long before the necessary 
probability distributions have been estimated reliably. The challenge is
usually not to implement an optimal system, given an infinite amount of
correctly labeled data, but to provide an acceptable performance on the
amount of training data that happens to be available.

\item[universality?] Are the unconditional probabilities really universal?
Is the notion of a population or ensemble valid enough to compute prior
probabilities of classes and evidences? Especially in language problems,
lack of universality becomes quickly evident: Contemporary corpora of
text, even if they encompass hundreds of thousands of documents, 
are utterly unusable
to help in deciphering a medieval text using statistical models. In order
to enjoy the benefits from computational linguistics, large reference corpora
of medieval documents in the same language and from a
comparable content would be needed, but these evidently do not exist.
The well-known large data set of {\em contemporary} handwritten digits MNIST 
with 50k samples~\citep{MNIST} is 
already inadequate for a reliable recognition of {\em late 19th/early 20th 
century} digits as we found out~\citep{bulacu2009recognition}.

\end{description}

\medskip

\section{Accuracy and quantization}

As a consequence of the problems noted here, textbook examples and
successful articles usually deal with questions of limited
dimensionality.  In a bioinformatics application with feature vectors of
20k dimensions, however, the product of probabilities for the observed
feature values would not often yield a meaningful value. This will
be dealt with in the next section. As an example:
 $0.999^{20000} = 2*10^{-9}$ but $0.998^{20000} = 0$ (at ANSI C double 
 or FORTRAN REAL*8 precision).  Sampling a
joint-probability distribution until reliable estimates are reached is
very expensive, even in problems of medium dimensionality.  Apart from
belief networks, the new field of conditional random fields (CRFs) is
now also focusing on Bayes as its computing element.  In the next
section we will show, that under conditions of unreliable probability
estimates, the chances of success are strongly limited for long chains
of probabilistic reasoning. 

\section{The product of probabilities}

An underestimated problem in Bayesian (and therefore Markovian)
modeling is the fact that a product of probabilities suffers from
a number of practical and fundamental limitations. Usually,
for a chain of observables or conditionals, a product of
probabilities is computed:

\begin{equation} \label{eq:prodprob1}
p_{true} = \prod_{i}^{n} p_{i} 
\end{equation}

where $n$ is the number of probabilities in the chain.
For example, for a number of class models $\Omega$, a model
is searched with the maximum probability product:

\begin{equation}
\Omega_{true} = argmax_{\Omega} \prod_{i}^{n} p_{i}(\Omega)
\end{equation}

What is usually forgotten is that each probability $p$ is
actually an estimate, most likely suffering from an error.
In the Bayesian belief approach, (initial) subjective probabilities
will be subject to such an estimation problem. However, also other
approaches entail an 'error in the probability', where that 
probability is estimated using, e.g., neural networks or other methods.
Let us assume there exists an additive error on the estimated
probabilities, such that a more realistic variation on Eq.~\ref{eq:prodprob1} is:

\begin{equation} \label{eq:prodprob2}
p_{measured}(t) = \prod_{i}^{n} \left(p_{i} + \epsilon(t)\right)
\end{equation}

where $\epsilon$ is some stochastic variable characterized by
a (usually unknown) statistical distribution. Then, the expected 
absolute-valued error is

\begin{equation} \label{eq:err1}
\varepsilon = E | p_{true} - p_{measured} |
\end{equation}

For sake of simplicity, in order to obtain an impression of
the order of magnitude of errors to be expected, assume
that all $p_{i} \approx P$, where P is a given constant
probability such that the product reduces to:

\begin{equation} \label{eq:simplepow}
p_{true} = P^n
\end{equation}

and

\begin{equation} \label{eq:powpluserr}
p_{measured} = (P+\epsilon)^n
\end{equation}

Then, the expected absolute-valued error can be expressed as:

\begin{equation} \label{eq:err2}
\varepsilon' = E | P^n - (P+\epsilon)^n |
\end{equation}

More interesting, however, is the relative error, as it
will immediately relate to the ranking of different models $\Omega$
in a classification problem.

\begin{equation} \label{eq:err3}
\varepsilon'_{rel} = \varepsilon' / P^n
\end{equation}

or

\begin{equation} \label{eq:err4}
\varepsilon'_{rel} \approx | P^n - (P+\epsilon)^n | / P^n 
\end{equation}

For non-zero $P$ we can now estimate the magnitude of
the relative error:

\begin{eqnarray}
\varepsilon'_{rel} & \approx & | 1 - (P+\epsilon)^n / P^n | \\
                & = & | 1 - \left(\frac{(P+\epsilon)}{P}\right)^n | \\
                & = & | 1 - \left(1+\epsilon/P\right)^n |
\end{eqnarray}

Only in the case $\epsilon$ is exactly equal to zero, the expected
relative error will equal zero. In all other cases, a power of
the sum of one plus an error fraction will lead to a very large
error for large $n$. For problems with a chain of probabilities for
more than ten pieces of (Bayesian) information, there will be a serious error. 

\medskip 

Additionally, but separately from the fundamental problem described here,
the product of probabilities entails very small values which lead
to quantization errors. In the programming language C, for instance, the use
of {\bf float} (4 bytes for a floating point number) must be excluded.
Regular {\bf double} precision (8 bytes) will do better, but it is advisable
to use {\bf long double}. A common trick is to sum $\log(p)$ instead of
computing a product. This may help somewhat in the computational
problem, but that does not solve the problem of an {\em error $\epsilon$ in
the probability estimates} biasing the final product of $p$ or sum of $log(p)$. 
In the next section an experiment is performed to check the predictions.

\subsection{Monte-Carlo experiment}

\begin{figure}[H] \center
\includegraphics[width=10cm]{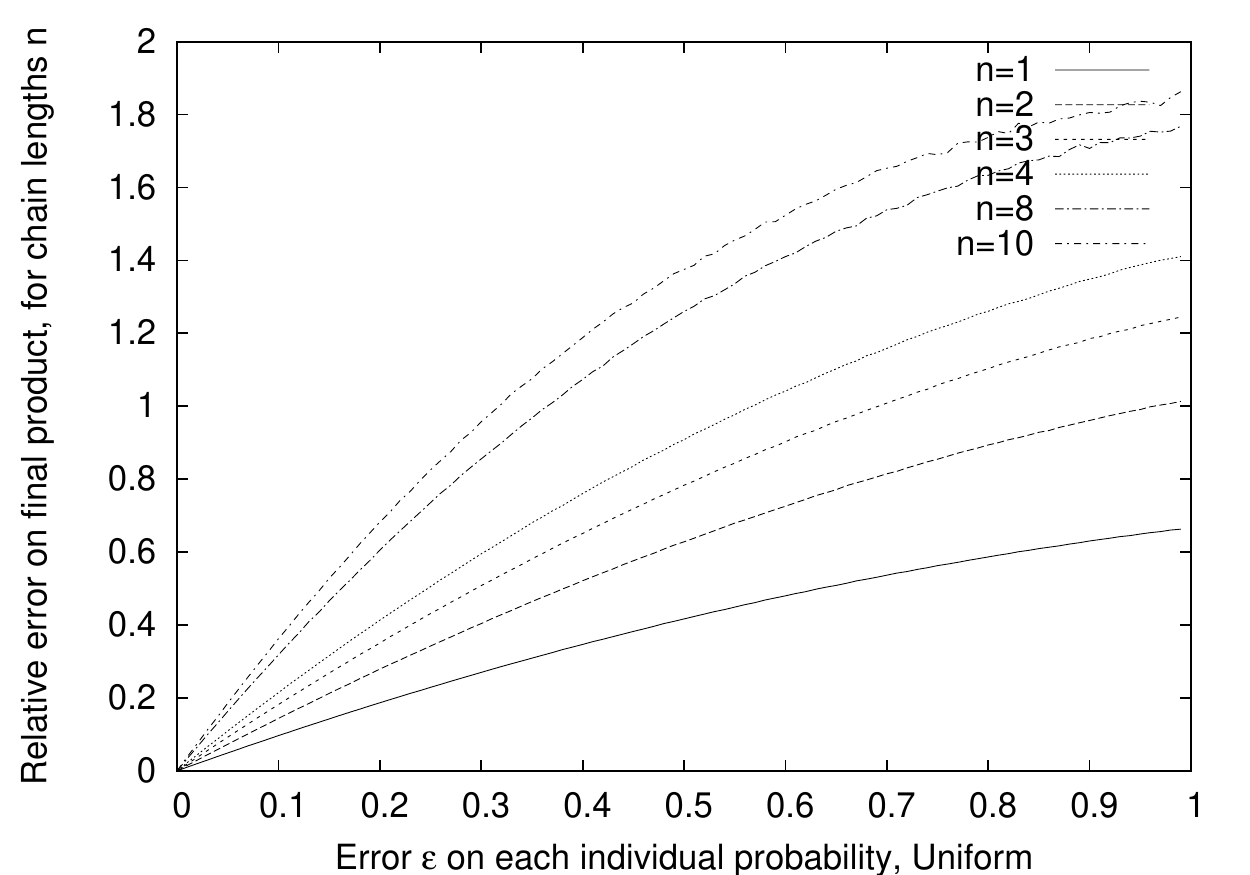}
\caption{\label{fig:curves1}Computed (experimental) relative errors on probability products for different error levels $\varepsilon$ and different chain lengths $n$. The x axis represents
the (input) error on the individual probabilities, expressed in numbers of standard deviations of a uniform distribution $[0,1]$: a value of 0.3
corresponds to 0.3 x $\sqrt{1/12}$ = 0.087.}
\end{figure}

\begin{figure}[H] \center
\includegraphics[width=10cm]{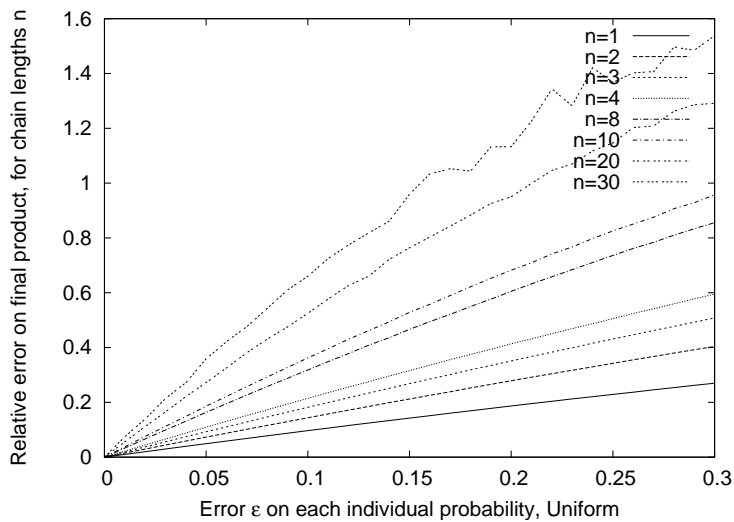}
\caption{\label{fig:curves2}Computed relative errors on probability products for different error levels $\varepsilon$ and different chain lengths $n$, for low to medium $\varepsilon$}
\end{figure}

\begin{figure}[H] \center
\includegraphics[width=10cm]{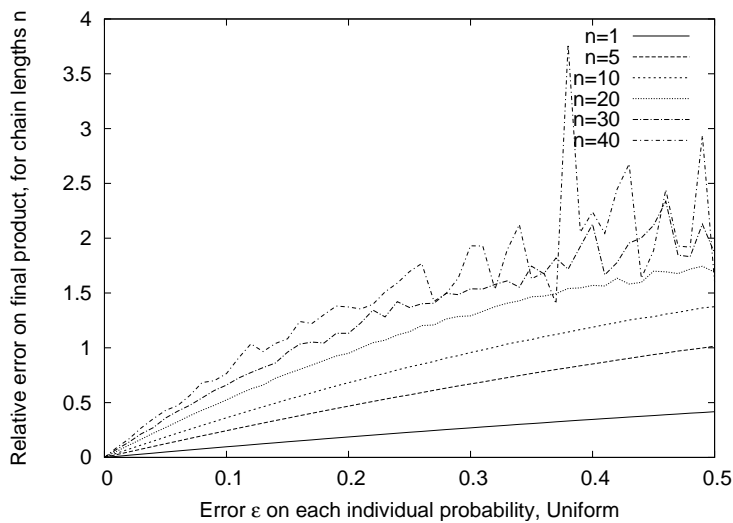}
\caption{\label{fig:curves3}Computed relative errors on probability products for different error levels $\varepsilon$ and different chain lengths $n$, for large $n$. 
Note the {\em ragged appearance} of the top curves with chain lengths $n=30$ and $40$. Even with 100k samples and 20 runs 
for averaging these curves, while using extended precision of the
floating-point values, there is a considerable quantization error, on top
of the {\em $\epsilon$-in-the-probability} phenomenon.}
\end{figure}

Apart from the analytical discussion above, it may be informative
to visualize the error propagation. A Monte-Carlo experiment is
performed. For a range of chain lengths $n$, for each value of $n$
a large number (100k) of probability vectors of dimensionality $n$ is
drawn (i.i.d.) with element values from a uniform distribution $x \in [0,1]$.
A range of additive-error amplitudes $a \in [0,1]$ 
is scanned in steps of 0.01, contaminating each true probability value with
a uniformly distributed $\epsilon \in [-a,a]$. For each instance of
a probability vector, the relative difference between the true probability product 
and the measured product -as contaminated by $\epsilon$- is computed. Since these
computations are highly influenced by quantization error, the simulation will be based on {\bf long
doubles} (16 bytes per floating-point value) to mitigate quantization error. 
Each experiment for a value 
of $a$ is repeated twenty times to obtain an average error curve (Figures~\ref{fig:curves1} to \ref{fig:curves3}).
With a chain of ten probabilities and an error amplitude $a = 0.3$ for $\epsilon$, the
resulting product of probabilities suffers from an error which is of
the same order of magnitude as the true probability product, i.e.,
the relative error has a value of one (100\%). For longer chains, the result
is evidently more dramatic. With a chain of 20 probabilities and the same
value of $\epsilon = 0.3$, the relative error is about 200\%!

These tests are rather primitive, in the sense that a uniform distribution
is used and the probability estimates are biased in a purely additive
fashion, without keeping the biased probabilities in the range $[0,1]$.
Both truncation and residing to a Gaussian error will mitigate the
error-multiplication effect. However, for a truncated uniform error (Appendix,
Table I), the case of ten probabilities in a chain will still 
lead to a relative error of about one (100\%). With a chain of twenty
probabilities, the relative error will be about 130\%.

For non-truncated Gaussian errors on the individual probabilities,
assuming a base standard deviation of $\sigma=\sqrt{1/12}$, which is the
same value as the standard deviation of the uniform distribution,
the effect is less extreme, but still highly worrying. Here, in case of a
chain of ten probabilities, the relative error will be 50\%. For a chain
of twenty probabilities, the relative error will be 80\% (Appendix, Table II).

In this section we have seen from Monte-Carlo experiments on products
of probabilities that the presence of an estimation error in the
chain will yields dramatic errors, even in problems of dimensionalities
that would be considered low (e.g., Ndim=20) in current-day problems
in scientific methods or applied systems.

\section{How is the error of a product of probabilities distributed?}

The performance of Monte-Carlo experiments leaves one with the unsatisfying
situation that a more compact analytical description of the problem
does not exist. According to the central limit theorem, the sum of arbitrary
distributions will converge to the Gaussian distribution. In case of 
the product of independent random variable, a log normal distribution
may be expected. However, it can be observed that not only the mean but
also the standard deviation of resulting distributions increases as chain lengths
will be longer. In this case, as yet without proof, we will propose
the continuous Poisson distribution.

When running a Monte-Carlo experiment
to draw $n$ probabilities $p_i$ from a uniform distribution
in order to multiply them, it will appear that with a large
enough number of iterations:

\begin{equation}
   \sum_{k=1}^{N} \sum_{i=1}^{n} ln(p_{i}) = -n
\end{equation}

where $N$ is the number of iterations indexed $k$.
This -maybe counterintuitive- result is due to the
following. The integral of $ln(x), x \in \mathbb{R}$ is given by:

\begin{eqnarray}
{\cal F}(x) & = & \int ln(x)dx \\
           & = & x \,\, ln(x) - x + C
\end{eqnarray}

For the interval $x \in [0,1]$ this yields the
following value:

\begin{eqnarray}
\int_{0}^{1}  ln(x)dx & = & {\cal F}(1) - {\cal F}(0) \\
  \                   & = &  1 \,\, ln(1) - 1 + C - (0 \,\, ln(0) + C) \\
  \                   & = &  -1
\end{eqnarray}

Using the property that for computing a sum, the inner and outer
loop may be swapped in the algorithm, and realizing that the sum of a 
large (infinite) number of uniform-random draws from the interval $[0,1]$ 
amounts to integrative sampling of that interval, we obtain the following
estimation of the mean of the resulting distribution:

\begin{eqnarray}
\sum_{i=1}^n \int_{0}^{1} ln(p_{i}) dp & = & n \int_{0}^{1} ln(x) dx \\
                       & = & -n 
\end{eqnarray}

Thus, chain length directly determines the average estimate 
of $\sum_{i=1}^n ln(p_i)$, in fact, as the mean $\mu = -n$, the
variance will be $\sigma^{2} = n$, again, provided that the
original random distribution is uniform. This distributions looks
like an exponential distribution at $n=1$, with a maximum on the
right and the tail on the left. For increasing values of $n$,
the distribution will move left, gradually becoming less skewed.

If $m = \lfloor -\sum_{i=1}^n ln(p_{i}) \rfloor $, where $m \in \mathbb{N}_{\geq0} $ is
then $m$ is distributed approximately according to Poisson:

\begin{equation}
  Poiss(m,n) = \frac{n^{m} e^{-m}}{m!}
\end{equation}

However, for a sum of log probabilities $x = -\sum_{i=1}^n ln(p_{i})$, 
a continuous random variable 
$x \in \mathbb{R}_{\geq0}$ is needed, 
unlike the discrete, countable stochastic variable underlying the
regular Poisson distribution, such that we need to assume
a variant of the Poisson distribution describing a continuous stochastic
variable. To this end, the gamma function $\Gamma()$ can be applied:

\begin{equation}
  cPoiss(x,n) = \frac{n^{x} e^{-x}}{\Gamma (x+1)}
\end{equation}

The continuous-Poisson function $cPoiss()$ will reliably model 
the distribution of a product 
of probabilities in the domain of a sum of log probabilities.

\begin{figure}[H] \center
\includegraphics[width=9cm]{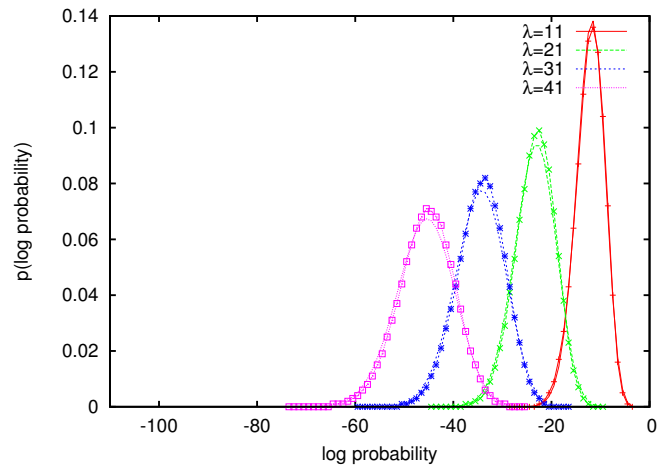}
\caption{\label{fig:model1}Computed (experimental, dotted line) and modeled distributions (solid thin line) 
of products of probabilities. The x axis represents the log probability 
value of a product chain. Eight curves are shown, for four 
values of the Poisson $\lambda$ (11,21,31,41) coinciding with chain lengths, probabilities were 
sampled in the range $p \in (0,0.84]$. The good overlap between experimental Monte-Carlo
results and the analytical Poisson curve makes distinction of these curves
difficult. Going from right to left: The longer the chain, the lower the
log probability, which is to be expected, but also the wider the 
dispersion of the error distribution, following the continuous Poisson
distribution.}
\end{figure}

\begin{figure}[H] \center
\includegraphics[width=9cm]{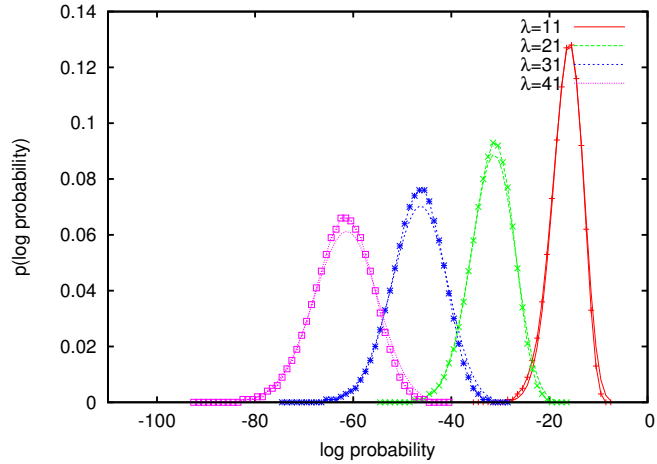}
\caption{\label{fig:model2}Computed (experimental) and modeled distributions 
of products of probabilities The x axis represents the log probability 
value of a product chain. Eight curves are shown, for four 
values of the Poisson $\lambda$ (11,21,31,41) coinciding with chain lengths, probabilities were 
sampled in the range $p \in (0,0.58]$.}
\end{figure}

Apart from a more or less practical premise which led us to propose the continuous
Poisson distribution, namely that a continuous variable requires a continuous
Poisson model, a formal proof is not given here. However, this model fits the
Monte-Carlo experiments very well. Figures~\ref{fig:model1} and ~\ref{fig:model2} show examples
of correspondence between data and continuous-Poisson model. The model requires
one scaling factor along the x-axis, to compensate for the simulation 
characteristics. The same value is used for all values of $n=\lambda$ in 
a graph.

The next question is how to model the distribution of a sum of the
logarithm of probabilities when the uniform distribution is in the domain 
$p \in [0,p_{max}]$ rather than in the interval $[0,1]$. For the case of
an arbitrary maximal probability $p_{max}$, the position of the continuous-Poisson 
distribution can be translated simply along the abscissa, by denoting $x' = x + ln(p_{max}^{n})$ 
and applying $cPoiss(x',n)$. Still further, a generalization can
be obtained by constraining the probabilities in an interval $[p_{min},p_{max}]$.
Under these conditions, with proper scaling, the continuous-Poisson function
will still be a good basis for modeling the effects of error transmission in
probability-product chains. More specifically, for a bipolar interval $[-e,e]$ 
and uniform noise, the standard deviation of the perturbations 
will be $\sigma_{e} = \sqrt{(2e)^2/12} = e/\sqrt{3}$
which will work out as a virtual elongation of the chain, such that the new
mean of a distribution of sum-log probabilities will be shifted leftwards from 
the value $n \, \ln{(p_{mean})}$ to the value $n \, \ln{(p_{mean}+\sigma_{e})}$.

\begin{table}[H]\centering
\caption{\label{tab:classifperf} F measure performance in percent, i.e., 
(precision+recall)/2, and standard deviation 
for the artificial word classification task, with
letter trees varying in depth and breadth, for varying values of the
additive random error bias $\epsilon$ on the letter transition probabilities of
given randomly drawn models. N=500 models per performance value. 
As an example, note the reduction in performance for word lengths of D=7 
from 93\% at $\epsilon=0.01$ to 85\% at $\epsilon=0.02$.}

\includegraphics[width=7cm]{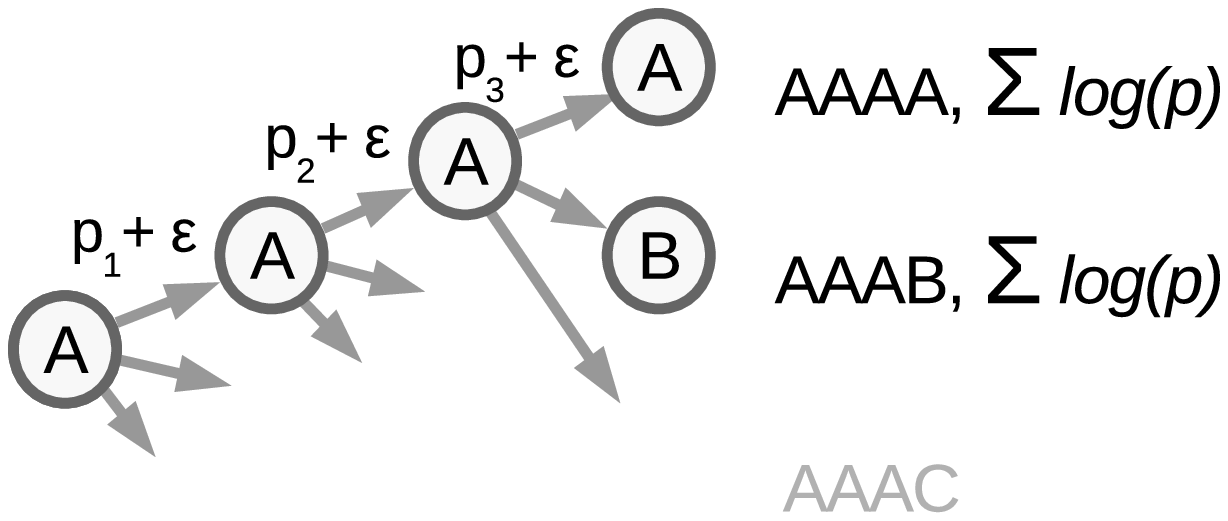}

\medskip
\small

\begin{tabular}{|l||rl||rl||rl||rl||rl||rl|} \hline
$\lambda$ \verb+\+ \hfill $\epsilon$ & 0.01 & sd & 0.02 & sd & 0.04 & sd & 0.08 & sd & 0.16 & sd & 0.32 & sd \\ \hline\hline
D=3 & 97.1 & {\scriptsize 8.8}& 93.6 & {\scriptsize 12.8}& 86.9 & {\scriptsize 17.8}& 74.2 & {\scriptsize 21.8}& 59.4 & {\scriptsize 22.4}& 35.8 & {\scriptsize 18.3}\\ 
D=5 & 95.0 & {\scriptsize 12.1}& 89.8 & {\scriptsize 16.7}& 82.1 & {\scriptsize 20.2}& 65.0 & {\scriptsize 24.5}& 43.7 & {\scriptsize 22.7}& 20.6 & {\scriptsize 15.4}\\ 
D=7 & 93.3 & {\scriptsize 13.0}& 85.2 & {\scriptsize 18.3}& 75.6 & {\scriptsize 22.0}& 55.7 & {\scriptsize 25.6}& 32.9 & {\scriptsize 21.9}& 12.1 & {\scriptsize 12.2}\\ 
D=9 & 91.6 & {\scriptsize 14.6}& 83.6 & {\scriptsize 18.7}& 69.0 & {\scriptsize 23.5}& 50.1 & {\scriptsize 25.2}& 26.5 & {\scriptsize 20.6}& 8.0 & {\scriptsize 9.5}\\ 
\hline 
{\em m} & {\em 94.2} & {\em\scriptsize 12.5}& {\em 88.1} & {\em\scriptsize 17.2}& {\em 78.4} & {\em\scriptsize 22.0}& {\em 61.2} & {\em\scriptsize 26.0}& {\em 40.6} & {\em\scriptsize 25.2}& {\em 19.1} & {\em\scriptsize 17.8}\\ \hline 
\end{tabular}

\end{table}

\section{Effects of 'error in the probability' on classification}

In order to test the effects of the error in the probability estimate
on a slightly more practical problem, consider the problem of finding
the path with the maximum sum of log(probability) in a synthetic lexical
search akin to word classification in a speech or handwriting recognition
system. The simulation is set up as follows. A balanced tree of given depth $d$
and breadth $b$ - i.e., the fan-out per node - is generated. Each node corresponds
with a letter of a word. Each edge is assigned a random transition probability
designated as the true probability. In the resulting tree, there will be a 
path of maximum product of probability, or maximum sum of log(probability), 
which is the target for classification. The other paths of the tree can be 
considered as distractors. Given this probabilistic structure, a number of 
test instances is created for a model, where each of the transition 
probabilities is nudged with a small additive uniform random error 
with zero mean and bipolar range $\pm\epsilon$. In the simulation, the following
parameters were varied: Depth $d={3,5,7,9}$, breadth $b={2,3,4,5,6}$ and
error  $\epsilon={0.01,0.02,0.04,0.08,0.16,0.32}$.
A wide range of models will thus be generated, randomly biased, and tested. 
For the performance of
the classification, the F measure will be used, with a weight of 0.5 for
precision and recall. Accuracy is not a good performance measure in this
test, because this performance measure is biased by a large number of correct 
rejects for distractor paths in the -possibly large- tree.

Table~\ref{tab:classifperf} shows the classification performance for
trees varying in depth from 3 to 13, breadth (fan out) varying from 2 to 6, 
and errors $\epsilon$ varying from 0.01 to 0.32, determining the amplitude
of bipolar pseudo-random uniform noise\footnote{drand48() was used}, with a 
mean of zero. The number of error-biasing trials for a given random model 
tree is 400. The number of randomly drawn models is 100 per tree topology. 
As can be inferred from the table, the expectation that errors will 
somehow be canceled out cannot be confirmed. Note that lexicon size
(8 to 279936 paths) and word length (3 to 7) were varied 
in order to compute average performances. For instance, one model test
for a tree with depth 7 and breadth 6 yields $6^7=279936$ different paths,
i.e., one target and 279935 distractor paths. This particular single model test
requires some $10^8$ sum of log-probability path evaluations and is 
repeated 100 times. Already
a probability error of only $\epsilon=0.01$ yields a drop of three percent
in path classification performance, even
on short paths. This deteriorates as paths become longer and the number
of distractors increases (Table~\ref{tab:classifperf}).

\section{Intermediate Conclusion: Error propagation in probability product chains}

In current applications in pattern recognition,
machine learning and data mining, feature vectors and problem dimensionality 
may well exceed an $n$ of ten dimensions. Dimensionalities of hundreds,
thousands, even tens of thousands (in bioinformatics and marketing) are
common. Therefore, it is highly likely that many systems and applications
reported in literature suffer from the problem described here. 
It is a riddle, how current text books keep advertising 
Bayesian and Markovian approaches, virtually without caveats. At the conceptual,
descriptive level, the reverend Bayes is right, of course. Who would want 
to argue with the basic principle?
For a realizable and powerful classification system one must hope that 
it {\em behaves} in a Bayesian manner, because this is the theoretical optimum. 

\medskip

The issue of one-trial (one-shot) learning has garnered attention in recent
years, both in the Bayesian~\citep{fei2006one,LakeConceptLearningScience} 
and in the deep-learning~\citep{OneShotDeepRezende2016} schools of thought.
Assuming that generally true, low-level patterns have been trained to a
system in a reliable manner, this 'world' information can be 
used in a generative fashion by making inferences on perceived instances of a 
new, previously unseen class. With the proper pretraining and dimension 
reduction, the existing Bayesian conditional probability landscape can be
used to infer new categories and classes. In no way this paper is intended to
question the importance of Bayesian modeling and the power of generative
approaches based on Bayes. However, generative hypothesis generation can
also be realized using deep learning, with a lesser stress on the Bayesian
properties ('approximate Bayesian inference'~\citep{Kingma2014}). It 
remains to be seen in how far the current results are still small and
specific, with little hope for easy replication in other domains.
If the product of probabilities chain can be kept short, 
the error propagation problem addressed in this paper may pose only a 
limited risk.

\medskip 

However, as demonstrated before, it is unrealistic to expect veridical a
posteriori probability estimates when computing in a Bayesian style over
a practical problem where many pieces of evidence are combined in a
product of probabilities.  An important motivation for writing this memo
is the regular stream of questions received from PhD students 
about HMMs for handwriting recognition because they encounter problems
in replicating good recognition result from HMM literature.  The reason
may be that failures are not archived in the public arena and 'typical
results' may indeed be very typical. 
\cite{vanOostenMarkov2014} addresses a number of 'tricks of the trade',
well known to researchers in the domain of hidden-Markov modeling
but not spelled out in enough detail in the literature. 


\section{Caveats on Markov modeling}

Although Hidden-Markov Models are well known, a brief introduction is necessary.
The assumption is that in stochastic observation series, the next observation
can predicted from the current state of a modeled system. For
explicit ('surface') Markov models, the current state for a discrete time and discrete state system can be the observation of a token from the alphabet of a stochastic
language generator. For large alphabets, such as the list of phonemes uttered by
a human speaker, it became quite clear, that in order to work for speech  recognition, the model space needed to be reduced: The number of
state transitions would consist of an L*L matrix, with L being the number of tokens
emitted by the source. Such a matrix can become quite large, because a typical
'alphabet' of quantized phonemes may have hundreds of tokens. A powerful solution for this problem is to replace the
concrete observation token-as-a-state by a more abstract, 'hidden', state, which induces a probability distribution of tokens that is valid when this state occurs.
In this manner, a number of $k<<L$ state transitions can be used to characterize
a state-dependent token-emission process. Also in the Markov paradigm, products of
probabilities play an essential role, for instance in the computation of the 
likelihood that a particular sequence of states represents, e.g., a spoken word
in speech recognition. Most commonly, sums of products of two probabilities are used of the form $p_{ij} = \sum_{r=1}^{k} p_{ir}p_{rj}$. This means that the propagation of the random estimation error $\epsilon$ within each value of $p$ 
is less dramatic than in a long belief-network reasoning chain but will still have an unavoidable problematic effect on the model estimation and its use for classification or prediction. 

A final caveat to be dealt with is the question whether the temporal structure
of a particular classification problem is important, at all. Does one really
need a state transition matrix for problems such as handwriting or speech 
recognition? Because these problems present themselves as temporal, with 
patterns of varying duration, Markov modeling seems to be unavoidable.

What is the word classification rate in a lexicon, if the only information used
is letter presence, regardless of sequential order of the letters? 
In order to test the uniqueness of words in lexica of a number of languages, 
an experiment was done, using nearest-neighbor classification. 
The feature vector for each word of a lexicon consists of the counts of its 
letters in the alphabet. The word 'a' would be represented by a vector with an
element of value one and 61 zeros (Upper and lower case letters are treated 
as different and international characters are transliterated to their
nearest ASCII representative, keeping the dimensionality identical for
all languages in the test). 
For nine languages, 20-thousand words were randomly 
selected each, from a larger lexicon. In order to prevent deterministic sort
order effects, a small fraction of noise was added to the counts, making sure
that the total variation remains within a step count of one (bipolar uniform
noise amplitude $a=0.0001$. In the worst case, $a * 62 << 1$, the deviation
from the letter count is much smaller than one, so no spurious attractors
for nearest-neighbor matching are created.
This was done twice per test per lexicon, such that a reference set and a
test set was obtained which marginally differed in terms of the small noise
perturbations\footnote{Lexical 'letvec' data can be obtained here: \url{monkweb.nl/monk/Projects/distro-UDI-Letvec.tgz}}.
In this manner, words with an identical letter count vector have an equal
probability of obtaining a hit in the nearest-neighbor search. 
Tests were repeated six times,
with random values. The random number generator was initialized by using the
microseconds field value of the time of program start.


\begin{table}\centering
\caption{\label{tab:letfreqs} Classification rate of finding a word
in a lexicon, based on its letter frequencies alone,
for lexica from a number of languages. Results are based on six 
randomly-selected 20k word sets per lexicon.}

\medskip

\begin{tabular}{|l|r|r|r|r|}  \hline
Language       & Lexicon size & Selected &   nearest neighbor & sd \\
\,             & (words)      & words/test & correct (\%) & (\%)\\ \hline \hline
Italian        &   62359           & 20k &         95.1 & 0.11 \\ \hline
Spanish        &   51568           & 20k &         95.6 & 0.19 \\ \hline
Portuguese     &   32451           & 20k &         96.2 & 0.17 \\ \hline
French         &   47618           & 20k &         97.1 & 0.18 \\ \hline
Danish         &   55485           & 20k &         97.8 & 0.08 \\ \hline
Dutch          &  114029           & 20k &         99.1 & 0.06 \\ \hline
UK-English     &   77389           & 20k &         98.6 & 0.09 \\ \hline
English (Unix) &   45402           & 20k &         97.3 & 0.09 \\ \hline
German         &   78228           & 20k &         99.6 & 0.04 \\ \hline
\end{tabular}
\end{table}
Table~\ref{tab:letfreqs} shows the results, which reveal a surprisingly
high recognition rate on the basis of word-letter frequency, with a stable
result over multiple random selections. On the basis of these findings,
speech and handwriting recognizers, which usually have lower classification
rates than this, should focus predominantly on letter-presence detection
rather than spending modeling time on the sequential order. Discussion
with colleagues from different teams in Markov modeling world-wide revealed that 
it was, paraphrasing, {\em 'considered a well-known fact that the 
transition matrix was less important than the state-conditional 
probability distributions in Markov modeling'}. According to these results,
the efforts in sequence modeling will only cover one to five percent of
the problems in lexical classification. It can be hypothesized that BLSTM
(neural) methods may also have relied more on pattern presence detection rather than
on elaborate sequence modeling, after training. In \cite{vanOostenMarkov2014} we
have shown that the importance of temporal modeling can easily be checked
when training a hidden Markov Model by flattening the transition matrix.
If the performance collapses, temporal modeling was essential. As in the 
lexical example above, the collapse may be limited since the essential
information for classification resides in the state-conditional observation
probability distributions of the HMM and not in the state transitions.

\section{Non-probabilistic distance-based matching}

\begin{figure}[H] \center
\includegraphics[width=10cm]{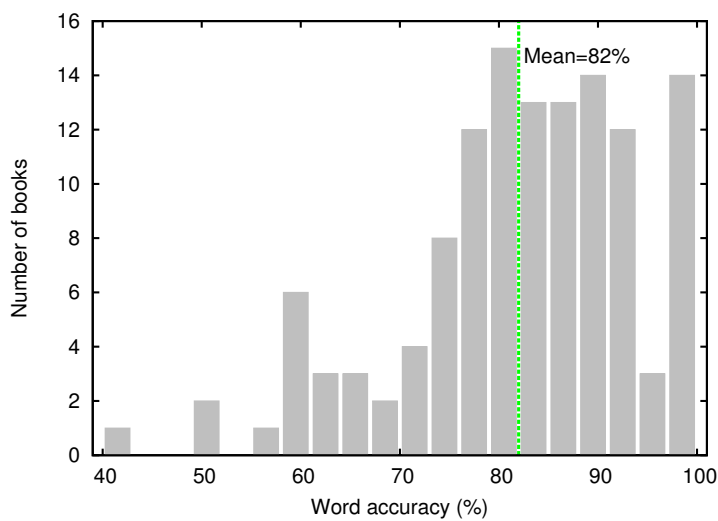}
\caption{\label{fig:myperfhisto}Histogram of word accuracy for 126 handwritten
documents in different scripts and languages, on words with 50 example instances, using a plain feature-based
approach and nearest-mean matching. Average lexicon size is 1100 words in
a document. Total number of word instances: 115940. In a continuous
learning setup, the difficult data sets on the left usually migrate
to the right as more ground-truth labels become available. In other words:
more than 50 samples per class appear to be needed for difficult material.}
\end{figure}

We have found that, using non-Markovian, non-Bayesian word classifiers
for historical handwriting in a wide range of scripts, periods and 
languages, nearest-centroid feature matching schemes may attain 
word classification rates of 82\%, for classes with 50 example instances
each, and lexica of 1100 words on average. The largest lexicon used
was for 22k Chinese characters. Such results were obtained without 
extensive training and efforts by human researchers. With more modeling 
effort, word recognition rates above 90\% were obtained in~\cite{vanZant2008}.
Figure~\ref{fig:myperfhisto} shows the distribution of nearest-neighbor
(nearest-mean) matching in our ongoing '24/7' learning search engine 
for handwritten words, the Monk project~\citep{SchomakerSearch2016}.
Note that in these methods, the performance relies more on geometry (shape) 
than on probability (familiarity) of patterns. The theory on similarity
spaces and multiple-instance learning~\citep{DuinDissim2013} has made great
strides forward in recent years and the focus is on geometry rather than
on probability, here also.

\section{Conclusions}

Notwithstanding current folklore, there are many problems associated
with the application of Bayesian learning and classification and
hidden-Markov modeling. The attractive, clear simplicity of Bayes
will appear to have been misleading as soon as the modeling deviates 
from bowls with colored marbles, i.e., a world of closed sets and 
countable events, to the real world of sets with unclear boundaries, 
errors in probability estimates, limited data and missing observations.
The use of probabilities should not be a goal in itself, if the final
decision criteria are based on an other domain, e.g., utility.
In computing products of probabilities or sums of log probabilities,
beware of the propagating (relative) error. The error appears to be 
considerable, seemingly according to a continuous Poisson distribution.

For hidden-Markov modeling, it is advisable to test whether temporal
modeling is essential, for a problem at hand. Flattening of 
the transition matrix can be used as a test to find out whether 
the Markov assumption is needed. If it is not, the problem may also be
cast as a simultaneous feature-processing classification of the pattern
as a whole. If the pattern-learning process is in a bootstrapping stage, 
with only limited numbers of examples such that the use of current 'deep
learning' is out of the question, purely geometric methods such as 
nearest-neighbor matching are still an attractive option, especially in
sparsely labeled 'big data'.


\acks{We would like to acknowledge support for this project
from the Dutch NWO (grants 612.066.514 and 640.002.402)}


\newpage
\vskip 0.2in
\bibliography{problems}

\newpage
\section{Appendix: Table I. Relative error on product of probabilities, for truncated-Uniform error on the individual probabilities, for chain lengths $n$}

 $e$ = amplitude of error, to be multiplied with the value of each sample from the
 uniform distribution with $\mu=0, \sigma=\sqrt{1/12}$.
 Truncation means that after the error addition, negative probabilities or values 
 above 1.0, are folded to zero or one, respectively.

\medskip

 $n$ = number of probabilities in the product chain

\medskip

 These values are computed experimentally and contain quantization errors,
 especially for high $n$. Example: for an absolute estimation error $e = 0.06$,
 the relative error on the product of probabilities will be 0.3 (30\%) in
 a chain of 13 probabilities in the product.

\small

\begin{verbatim}

e \  n 1   3   5   7   9  11  13  15  17  19  21  23  25  27  29  31  33  35  37  39

0.00  0.0 0.0 0.0 0.0 0.0 0.0 0.0 0.0 0.0 0.0 0.0 0.0 0.0 0.0 0.0 0.0 0.0 0.0 0.0 0.0
0.01  0.0 0.0 0.0 0.0 0.0 0.0 0.0 0.0 0.1 0.1 0.1 0.1 0.1 0.1 0.1 0.1 0.1 0.1 0.1 0.1
0.02  0.0 0.0 0.1 0.1 0.1 0.1 0.1 0.1 0.1 0.1 0.1 0.1 0.1 0.1 0.1 0.1 0.2 0.2 0.2 0.2
0.03  0.0 0.1 0.1 0.1 0.1 0.1 0.1 0.1 0.2 0.2 0.2 0.2 0.2 0.2 0.2 0.2 0.2 0.2 0.2 0.3
0.04  0.0 0.1 0.1 0.1 0.1 0.2 0.2 0.2 0.2 0.2 0.2 0.2 0.2 0.3 0.3 0.3 0.3 0.3 0.3 0.3
0.05  0.0 0.1 0.1 0.2 0.2 0.2 0.2 0.2 0.3 0.3 0.3 0.3 0.3 0.3 0.3 0.4 0.4 0.4 0.4 0.4
0.06  0.1 0.1 0.1 0.2 0.2 0.2 0.3 0.3 0.3 0.3 0.3 0.4 0.4 0.4 0.4 0.4 0.5 0.5 0.5 0.5
0.07  0.1 0.1 0.2 0.2 0.2 0.3 0.3 0.3 0.3 0.4 0.4 0.4 0.4 0.4 0.5 0.5 0.5 0.5 0.6 0.6
0.08  0.1 0.1 0.2 0.2 0.3 0.3 0.3 0.4 0.4 0.4 0.4 0.5 0.5 0.5 0.6 0.6 0.6 0.6 0.6 0.7
0.09  0.1 0.2 0.2 0.3 0.3 0.3 0.4 0.4 0.4 0.5 0.5 0.5 0.5 0.6 0.5 0.6 0.6 0.7 0.7 0.7
0.10  0.1 0.2 0.2 0.3 0.3 0.4 0.4 0.5 0.5 0.5 0.5 0.6 0.6 0.6 0.7 0.7 0.7 0.8 0.7 0.8
0.11  0.1 0.2 0.3 0.3 0.4 0.4 0.5 0.5 0.5 0.6 0.6 0.6 0.6 0.7 0.7 0.7 0.7 0.8 0.9 0.8
0.12  0.1 0.2 0.3 0.4 0.4 0.5 0.5 0.5 0.6 0.6 0.6 0.7 0.7 0.7 0.8 0.8 0.8 0.9 1.0 0.9
0.13  0.1 0.2 0.3 0.4 0.4 0.5 0.5 0.6 0.6 0.6 0.7 0.7 0.8 0.8 0.8 0.9 0.9 0.9 1.0 0.9
0.14  0.1 0.3 0.3 0.4 0.5 0.5 0.6 0.6 0.7 0.7 0.7 0.8 0.8 0.8 0.9 0.9 1.0 0.9 1.0 1.0
0.15  0.1 0.3 0.4 0.4 0.5 0.6 0.6 0.7 0.7 0.7 0.8 0.8 0.9 0.9 0.9 0.9 1.0 1.0 1.0 1.0
0.16  0.2 0.3 0.4 0.5 0.5 0.6 0.6 0.7 0.7 0.8 0.8 0.8 0.9 0.9 0.9 1.0 1.1 1.1 1.1 1.2
0.17  0.2 0.3 0.4 0.5 0.6 0.6 0.7 0.7 0.8 0.8 0.9 0.9 1.0 1.0 1.0 1.0 1.1 1.2 1.1 1.1
0.18  0.2 0.3 0.4 0.5 0.6 0.7 0.7 0.8 0.8 0.9 0.9 1.0 1.0 1.0 1.1 1.1 1.1 1.2 1.2 1.7
0.19  0.2 0.3 0.4 0.5 0.6 0.7 0.7 0.8 0.9 0.9 1.0 1.0 1.0 1.1 1.1 1.2 1.1 1.3 1.3 1.4
0.20  0.2 0.4 0.5 0.6 0.6 0.7 0.8 0.8 0.9 0.9 1.0 1.0 1.1 1.1 1.1 1.2 1.2 1.3 1.2 1.6
0.21  0.2 0.4 0.5 0.6 0.7 0.7 0.8 0.9 0.9 1.0 1.0 1.1 1.1 1.2 1.2 1.2 1.3 1.3 1.3 1.3
0.22  0.2 0.4 0.5 0.6 0.7 0.8 0.8 0.9 1.0 1.0 1.1 1.1 1.1 1.2 1.3 1.3 1.3 1.2 1.6 1.4
0.23  0.2 0.4 0.5 0.6 0.7 0.8 0.9 0.9 1.0 1.1 1.1 1.2 1.2 1.3 1.2 1.3 1.4 1.4 1.5 1.9
0.24  0.2 0.4 0.6 0.7 0.8 0.8 0.9 1.0 1.0 1.1 1.1 1.2 1.2 1.3 1.6 1.4 1.5 1.3 1.4 1.4
0.25  0.2 0.4 0.6 0.7 0.8 0.9 0.9 1.0 1.1 1.1 1.2 1.2 1.3 1.4 1.4 1.4 1.4 1.4 1.4 1.5
0.26  0.2 0.4 0.6 0.7 0.8 0.9 1.0 1.0 1.1 1.2 1.2 1.3 1.3 1.4 1.4 1.4 1.6 1.6 1.7 1.5
0.27  0.2 0.5 0.6 0.7 0.8 0.9 1.0 1.1 1.1 1.2 1.2 1.3 1.3 1.4 1.3 1.4 1.5 1.5 1.5 1.7
0.28  0.3 0.5 0.6 0.8 0.9 0.9 1.0 1.1 1.2 1.2 1.3 1.3 1.4 1.4 1.4 1.4 1.6 1.4 1.3 1.6
0.29  0.3 0.5 0.7 0.8 0.9 1.0 1.1 1.1 1.2 1.3 1.3 1.3 1.4 1.5 1.5 1.4 1.4 1.5 1.4 1.8
0.30  0.3 0.5 0.7 0.8 0.9 1.0 1.1 1.2 1.2 1.3 1.3 1.4 1.4 1.5 1.5 1.5 1.6 1.6 1.5 1.7

\end{verbatim}
\newpage
\section{Appendix: Table II. Relative error on product of probabilities, for non-truncated Gaussian error on the individual probabilities, for chain lengths $n$}

 $e$ = amplitude of error, to be multiplied with sample from a 
 normal distribution with $\mu=0, \sigma=\sqrt{1/12}$ which are
 the same values as in the tests on the uniform distribution.

\medskip

 $n$ = number of probabilities in the product chain

\medskip

 These values are computed experimentally and contain quantization errors,
 especially for high $n$. Example: for an absolute estimation error $e = 0.07$,
 the relative error on the product of probabilities will be 0.2 (20\%) in
 a chain of 13 Gauss-distributed probabilities in the product.

\small

\begin{verbatim}

e \  n 1   3   5   7   9  11  13  15  17  19  21  23  25  27  29  31  33  35  37  39

0.00  0.0 0.0 0.0 0.0 0.0 0.0 0.0 0.0 0.0 0.0 0.0 0.0 0.0 0.0 0.0 0.0 0.0 0.0 0.0 0.0
0.01  0.0 0.0 0.0 0.0 0.0 0.0 0.0 0.0 0.0 0.0 0.0 0.0 0.0 0.0 0.0 0.0 0.0 0.0 0.0 0.0
0.02  0.0 0.0 0.0 0.0 0.0 0.0 0.0 0.0 0.1 0.1 0.1 0.1 0.1 0.1 0.1 0.1 0.1 0.1 0.1 0.1
0.03  0.0 0.0 0.0 0.0 0.1 0.1 0.1 0.1 0.1 0.1 0.1 0.1 0.1 0.1 0.1 0.1 0.1 0.1 0.1 0.1
0.04  0.0 0.0 0.0 0.1 0.1 0.1 0.1 0.1 0.1 0.1 0.1 0.1 0.1 0.1 0.2 0.2 0.2 0.2 0.2 0.2
0.05  0.0 0.0 0.1 0.1 0.1 0.1 0.1 0.1 0.1 0.1 0.1 0.1 0.2 0.2 0.2 0.2 0.2 0.2 0.2 0.2
0.06  0.0 0.1 0.1 0.1 0.1 0.1 0.1 0.1 0.2 0.2 0.2 0.2 0.2 0.2 0.2 0.2 0.2 0.3 0.2 0.3
0.07  0.0 0.1 0.1 0.1 0.1 0.1 0.2 0.2 0.2 0.2 0.2 0.2 0.2 0.3 0.3 0.3 0.3 0.3 0.3 0.3
0.08  0.0 0.1 0.1 0.1 0.1 0.2 0.2 0.2 0.2 0.2 0.2 0.3 0.3 0.3 0.3 0.3 0.3 0.3 0.4 0.4
0.09  0.0 0.1 0.1 0.1 0.2 0.2 0.2 0.2 0.2 0.2 0.3 0.3 0.3 0.3 0.3 0.3 0.3 0.3 0.4 0.4
0.10  0.0 0.1 0.1 0.1 0.2 0.2 0.2 0.2 0.3 0.3 0.3 0.3 0.3 0.3 0.4 0.4 0.4 0.5 0.5 0.4
0.11  0.1 0.1 0.1 0.2 0.2 0.2 0.2 0.3 0.3 0.3 0.3 0.3 0.4 0.4 0.4 0.4 0.5 0.5 0.4 0.5
0.12  0.1 0.1 0.1 0.2 0.2 0.2 0.3 0.3 0.3 0.3 0.3 0.4 0.4 0.4 0.5 0.4 0.5 0.5 0.5 0.6
0.13  0.1 0.1 0.2 0.2 0.2 0.3 0.3 0.3 0.3 0.3 0.4 0.4 0.4 0.4 0.5 0.5 0.5 0.5 0.6 0.6
0.14  0.1 0.1 0.2 0.2 0.2 0.3 0.3 0.3 0.3 0.4 0.4 0.4 0.4 0.5 0.5 0.5 0.6 0.6 0.7 0.6
0.15  0.1 0.1 0.2 0.2 0.3 0.3 0.3 0.4 0.4 0.4 0.5 0.5 0.5 0.5 0.5 0.5 0.6 0.7 0.6 0.7
0.16  0.1 0.1 0.2 0.2 0.3 0.3 0.4 0.4 0.4 0.4 0.5 0.5 0.5 0.6 0.6 0.6 0.7 0.7 0.8 0.7
0.17  0.1 0.2 0.2 0.3 0.3 0.3 0.4 0.4 0.4 0.5 0.5 0.5 0.6 0.6 0.7 0.7 0.7 0.8 0.7 0.8
0.18  0.1 0.2 0.2 0.3 0.3 0.4 0.4 0.4 0.5 0.5 0.5 0.6 0.6 0.6 0.6 0.7 0.7 0.7 0.8 0.9
0.19  0.1 0.2 0.2 0.3 0.3 0.4 0.4 0.5 0.5 0.5 0.6 0.6 0.6 0.7 0.6 0.8 0.8 0.8 1.0 1.1
0.20  0.1 0.2 0.2 0.3 0.4 0.4 0.4 0.5 0.5 0.5 0.6 0.6 0.7 0.7 0.8 0.8 0.9 0.9 1.0 0.7
0.21  0.1 0.2 0.3 0.3 0.4 0.4 0.5 0.5 0.5 0.6 0.6 0.7 0.7 0.8 0.7 0.9 0.9 0.9 1.0 0.9
0.22  0.1 0.2 0.3 0.3 0.4 0.4 0.5 0.5 0.6 0.6 0.6 0.7 0.7 0.8 0.8 0.8 0.9 0.9 1.2 1.1
0.23  0.1 0.2 0.3 0.3 0.4 0.5 0.5 0.6 0.6 0.6 0.7 0.7 0.8 0.8 0.8 1.0 0.8 0.9 0.9 1.1
0.24  0.1 0.2 0.3 0.4 0.4 0.5 0.5 0.6 0.6 0.7 0.7 0.7 0.8 0.9 0.9 0.9 1.1 1.0 0.9 1.1
0.25  0.1 0.2 0.3 0.4 0.4 0.5 0.5 0.6 0.6 0.7 0.7 0.8 0.8 0.9 1.0 1.2 1.0 1.0 1.1 1.2
0.26  0.1 0.2 0.3 0.4 0.5 0.5 0.6 0.6 0.7 0.7 0.8 0.8 0.8 0.9 1.2 1.1 1.1 1.2 1.2 1.2
0.27  0.1 0.2 0.3 0.4 0.5 0.5 0.6 0.6 0.7 0.7 0.8 0.9 0.9 1.0 1.1 1.0 1.1 1.1 1.3 1.2
0.28  0.1 0.3 0.3 0.4 0.5 0.6 0.6 0.7 0.7 0.8 0.8 0.9 0.9 1.0 1.1 1.0 1.1 1.1 1.4 1.4
0.29  0.1 0.3 0.4 0.4 0.5 0.6 0.6 0.7 0.7 0.8 0.9 1.0 0.9 0.9 1.2 1.2 1.1 1.1 1.1 1.3
0.30  0.1 0.3 0.4 0.5 0.5 0.6 0.7 0.7 0.8 0.8 0.9 0.9 1.0 1.0 1.2 1.3 1.3 1.3 1.5 1.5
\end{verbatim}
\newpage
\section{Appendix: Table III. Relative error on product of probabilities, for truncated-Gaussian error on the individual probabilities, for chain lengths $n$}

 $e$ = amplitude of error, to be multiplied with sample from a 
 normal distribution with $\mu=0, \sigma=\sqrt{1/12}$ which are
 the same values as in the tests on the uniform distribution.
Truncation means that negative probabilities or values above 1.0 are folded to
 either zero or one, respectively.
 
\medskip

 $n$ = number of probabilities in the product chain

\medskip

 These values are computed experimentally and contain quantization errors,
 especially for high $n$.

\small

\begin{verbatim}

e \  n 1   3   5   7   9  11  13  15  17  19  21  23  25  27  29  31  33  35  37  39

0.00  0.0 0.0 0.0 0.0 0.0 0.0 0.0 0.0 0.0 0.0 0.0 0.0 0.0 0.0 0.0 0.0 0.0 0.0 0.0 0.0
0.01  0.0 0.0 0.0 0.0 0.0 0.0 0.0 0.0 0.0 0.0 0.0 0.0 0.0 0.0 0.0 0.0 0.0 0.0 0.0 0.0
0.02  0.0 0.0 0.0 0.0 0.0 0.0 0.0 0.0 0.1 0.1 0.1 0.1 0.1 0.1 0.1 0.1 0.1 0.1 0.1 0.1
0.03  0.0 0.0 0.0 0.0 0.1 0.1 0.1 0.1 0.1 0.1 0.1 0.1 0.1 0.1 0.1 0.1 0.1 0.1 0.1 0.1
0.04  0.0 0.0 0.0 0.1 0.1 0.1 0.1 0.1 0.1 0.1 0.1 0.1 0.1 0.1 0.1 0.1 0.1 0.2 0.2 0.2
0.05  0.0 0.0 0.1 0.1 0.1 0.1 0.1 0.1 0.1 0.1 0.1 0.2 0.2 0.2 0.2 0.2 0.2 0.2 0.2 0.2
0.06  0.0 0.1 0.1 0.1 0.1 0.1 0.1 0.1 0.1 0.2 0.2 0.2 0.2 0.2 0.2 0.2 0.3 0.2 0.2 0.2
0.07  0.0 0.1 0.1 0.1 0.1 0.1 0.1 0.2 0.2 0.2 0.2 0.2 0.2 0.2 0.3 0.3 0.3 0.3 0.3 0.3
0.08  0.0 0.1 0.1 0.1 0.1 0.2 0.2 0.2 0.2 0.2 0.2 0.2 0.2 0.3 0.3 0.3 0.3 0.3 0.3 0.3
0.09  0.0 0.1 0.1 0.1 0.2 0.2 0.2 0.2 0.2 0.2 0.3 0.3 0.3 0.3 0.3 0.4 0.3 0.4 0.4 0.4
0.10  0.0 0.1 0.1 0.1 0.2 0.2 0.2 0.2 0.3 0.3 0.3 0.3 0.3 0.3 0.3 0.3 0.4 0.4 0.5 0.5
0.11  0.0 0.1 0.1 0.2 0.2 0.2 0.2 0.2 0.3 0.3 0.3 0.3 0.3 0.4 0.4 0.4 0.4 0.5 0.5 0.5
0.12  0.1 0.1 0.1 0.2 0.2 0.2 0.2 0.3 0.3 0.3 0.3 0.3 0.4 0.4 0.4 0.4 0.5 0.4 0.5 0.5
0.13  0.1 0.1 0.2 0.2 0.2 0.2 0.3 0.3 0.3 0.3 0.4 0.4 0.4 0.4 0.5 0.5 0.5 0.5 0.5 0.6
0.14  0.1 0.1 0.2 0.2 0.2 0.3 0.3 0.3 0.3 0.4 0.4 0.4 0.4 0.4 0.5 0.5 0.5 0.5 0.6 0.5
0.15  0.1 0.1 0.2 0.2 0.3 0.3 0.3 0.3 0.4 0.4 0.4 0.4 0.5 0.5 0.5 0.5 0.6 0.6 0.6 0.7
0.16  0.1 0.1 0.2 0.2 0.3 0.3 0.3 0.4 0.4 0.4 0.4 0.4 0.5 0.5 0.5 0.6 0.6 0.7 0.6 0.7
0.17  0.1 0.1 0.2 0.2 0.3 0.3 0.3 0.4 0.4 0.4 0.5 0.5 0.5 0.6 0.6 0.6 0.6 0.7 0.8 0.7
0.18  0.1 0.2 0.2 0.3 0.3 0.3 0.4 0.4 0.4 0.4 0.5 0.5 0.5 0.6 0.6 0.6 0.6 0.6 0.7 0.7
0.19  0.1 0.2 0.2 0.3 0.3 0.4 0.4 0.4 0.5 0.5 0.5 0.5 0.6 0.6 0.6 0.7 0.7 0.7 0.7 0.8
0.20  0.1 0.2 0.2 0.3 0.3 0.4 0.4 0.4 0.5 0.5 0.5 0.6 0.6 0.6 0.7 0.7 0.8 0.8 0.8 0.8
0.21  0.1 0.2 0.2 0.3 0.3 0.4 0.4 0.5 0.5 0.5 0.6 0.6 0.6 0.6 0.7 0.7 0.7 0.8 0.7 0.9
0.22  0.1 0.2 0.3 0.3 0.4 0.4 0.4 0.5 0.5 0.6 0.6 0.6 0.7 0.7 0.7 0.9 0.8 0.8 0.9 1.0
0.23  0.1 0.2 0.3 0.3 0.4 0.4 0.5 0.5 0.5 0.6 0.6 0.7 0.6 0.7 0.8 0.8 0.8 0.8 0.9 0.9
0.24  0.1 0.2 0.3 0.3 0.4 0.4 0.5 0.5 0.6 0.6 0.6 0.7 0.7 0.8 0.8 0.8 0.9 0.8 0.9 1.0
0.25  0.1 0.2 0.3 0.4 0.4 0.5 0.5 0.5 0.6 0.6 0.6 0.6 0.7 0.8 0.8 0.8 0.8 0.9 1.0 1.1
0.26  0.1 0.2 0.3 0.4 0.4 0.5 0.5 0.6 0.6 0.6 0.7 0.7 0.7 0.9 0.8 0.9 0.9 0.9 1.1 1.1
0.27  0.1 0.2 0.3 0.4 0.4 0.5 0.5 0.6 0.6 0.7 0.7 0.7 0.8 0.8 0.9 0.9 1.0 1.0 1.0 0.9
0.28  0.1 0.2 0.3 0.4 0.4 0.5 0.6 0.6 0.7 0.7 0.7 0.7 0.8 0.9 0.9 1.1 0.9 1.0 1.1 1.1
0.29  0.1 0.2 0.3 0.4 0.5 0.5 0.6 0.6 0.6 0.7 0.7 0.8 0.8 0.9 0.9 0.9 0.9 1.0 1.0 1.0
0.30  0.1 0.3 0.3 0.4 0.5 0.5 0.6 0.6 0.7 0.7 0.8 0.8 0.8 0.9 0.9 1.1 1.0 1.0 1.0 1.1

\end{verbatim}
\newpage
\section{Appendix: C Source for generating random probability sequences with 
                     an error in each estimate}

\tiny

\begin{verbatim}


 /* mulprob - Lambert Schomaker - May 2009 */

#include <stdio.h>
#include <stdlib.h>
#include <math.h>
#define N 2000
#define Nerr 100

int main(int argc, char *argv[])
{
   int i,j,k,n = N, niter=100000, seed, icurve, ncurves = 20;
   long double p[N], a, da = 0.01, e, d, sumd, sum, prod, target;
   double xax[Nerr], curve[Nerr];
   
   n = atoi(argv[1]);
   seed = atoi(argv[2]);
   srand48(seed);
   
   /* Clear arrays */
   
   a = 0.0;
   for(k = 0;  k < Nerr; ++k) {
      xax[k] = a;
      curve[k] = 0.0;
      a += da;
   }
   
   for(icurve = 0; icurve < ncurves; ++icurve) {
   
   /* For a number of error levels a (i.e., error on the p[]) */
   
      a = 0.0;
      for(k = 0; k < Nerr; ++k) {
         
        /* For a number of iterations to get a reliable estimate */ 
       
         sum = 0.0;
         sumd = 0.0;
         for(i = 0; i < niter; ++i) {
            
            /* Generate a sample of probabilities */ 
             
            for(j = 0; j < n; ++j) {
               p[j] = drand48();
            }
            
            /* Compute their veridical product */
         
            prod = 1.0;
            for(j = 0; j < n; ++j) {
               prod = prod * p[j]; 
            }
            target = prod;
            
            /* Compute their product, after adding bipolar noise */
         
            prod = 1.0;
   
            for(j = 0; j < n; ++j) {
               e = 2. * a * (drand48()-0.5);
               prod = prod * (p[j] + e); 
            }
            d = (target - prod);
            if(d < 0.) d = -d;
            sumd += d;
            sum += target;
         }
         curve[k] += (double) (sumd/sum);
         a += da;
      }
   } /* next curve */
   
   for(k = 0; k < Nerr; ++k) {
      curve[k] /= ncurves;
      printf("%f %f\n", xax[k], curve[k]);
   }
   return(0);
}
\end{verbatim}

\end{document}